\documentclass[manuscript,screen,sigconf]{acmart}

\AtBeginDocument{%
  \providecommand\BibTeX{{%
    \normalfont B\kern-0.5em{\scshape i\kern-0.25em b}\kern-0.8em\TeX}}}

\renewcommand\footnotetextcopyrightpermission[1]{}

\usepackage{multirow}
\usepackage{graphicx}
\usepackage{algorithm}
\usepackage{algorithmic}
\usepackage{flushend}
\usepackage{color}
\usepackage{mathtools}
\usepackage{tabularx}
\usepackage{subfig}
\usepackage{balance}
\usepackage{textcomp}
\usepackage{xcolor}

\begin{document}
\title{Vietnamese Poem Generation \& The Prospect Of Cross-Language Poem-To-Poem Translation}

\author{Triet Minh Huynh}
\orcid{0009-0009-2466-7022}
\affiliation{%
  \institution{Dept. of Information Technology Specialization, FPT University}
  \city{Ho Chi Minh City}
  \country{Vietnam}
}
\email{triethmse160251@fpt.edu.vn}

\author{Quan Le Bao}
\orcid{0009-0003-5651-7877}
\affiliation{%
  \institution{Dept. of Information Technology Specialization, FPT University}
  \city{Ho Chi Minh City}
  \country{Vietnam}
}
\email{quanlbse160758@fpt.edu.vn}


\begin{abstract}
Poetry generation has been a challenging task in the field of Natural Language Processing, as it requires the model to understand the nuances of language, sentiment, and style. In this paper, we propose using Large Language Models to generate Vietnamese poems of various genres from natural language prompts, thereby facilitating an intuitive process with enhanced content control. Our most efficacious model, the GPT-3 Babbage variant, achieves a custom evaluation score of 0.8, specifically tailored to the "luc bat" genre of Vietnamese poetry. Furthermore, we also explore the idea of paraphrasing poems into normal text prompts and yield a relatively high score of 0.781 in the "luc bat" genre. This experiment presents the potential for cross-Language poem-to-poem translation with translated poems as the inputs while concurrently maintaining complete control over the generated content.
\end{abstract}



\maketitle
\pagestyle{plain} 

\section{Introduction}

The rapid evolution of transformers \cite{vaswani2023attention} in the field of natural language processing has sparked our team's awareness of an unexplored frontier within its creative domain—specifically, the untapped potential of prompt-based poetry generation.

Existing implementations of poetry generation exhibit notable limitations in terms of input flexibility, often relying on a sparse set of initial keywords with minimal guidance for the poem's body. Surprisingly, there is no prominent model trained exclusively for Vietnamese poems utilizing the capabilities of GPT-3 (Generative Pretrained Transformers) \cite{brown2020language} and other large language models (LLMs).

Motivated by this void, our initiative aims to construct two architectures capable of accommodating natural language inputs, including instructional prompts specifying themes, styles, or content using GPT-3 and BLOOM \cite{unknown}. The envisioned outcome is a poem that is not only creative and unique, but also imbued with the intended sentiment and follows strictly the various rigid rules of poetry.

In the subsequent section, we will provide a brief overview of pertinent literature concerning transformers and poetry creation. This discussion will delve into their respective approaches, achievements, and inherent limitations.

\section{Related works}

Early attempts at poem generation often encountered the critical challenge of shallow content comprehension. These models relied heavily on pre-defined generation templates \cite{inproceedings}, term-based retrieval and reorganization to fit tonal and rhyme requirements \cite{inproceedings2}, or translation-based statistical approach for picking the most probable following sentence \cite{inproceedings3}, resulting in poems lacking creativity, coherency and controllability. Recent years have witnessed significant advancements in this field, evident in models utilizing recurrent neural networks (RNNs) like those proposed by Wang et al. \cite{wang2016chinese}, as well as the application of attention mechanisms \cite{wang2016chinese2, 9207442}. These approaches have demonstrably enhanced the fluency and meaningfulness of generated poems. While Tuan Nguyen et al. \cite{nguyen2021spgpt2} stands as the sole recent contribution to Vietnamese poetry generation, utilizing GPT-2 as the core foundation, they are still limited to predefined themes and starting keywords. 

But eversince the impressive breakthroughs achieved by large language models like OpenAI's ChatGPT (GPT-3.5) and GPT-3 \cite{brown2020language}, the field of NLP has been propelled to the forefront of public attention. Notably, ChatGPT's remarkable skill in comprehending the intricacies and abstractions of casual language \cite{ye2023comprehensive, sawicki2023bits} presents a compelling opportunity to revisit the domain of complex linguistic compositions such as poetry. This paper proposes a prompt-based approach that integrates the capabilities of LLMs to push the boundaries of Vietnamese poetry generation.

\section{Dataset}
In this section, we will provide an overview of the dataset utilized in our study, coupled with an exploration of the custom evaluation functions tailored for it.

Tuan Nguyen et al. \cite{nguyen2021spgpt2} have already released their dataset of 171,188 poems across five distinct genres, namely "4 chu", "5 chu", "7 chu", "luc bat", and "8 chu", with "luc bat" genre being the most popular with 87,609 samples. Subsequent to the acquisition of this dataset, a process of filtering was conducted to identify poems deemed of superior quality, facilitated through the establishment of a scoring system. This scoring system, inspired by the evaluation algorithm employed by Tuan Nguyen et al., has been extended to encompass all genres and serves a dual purpose in both the filtration of poems and the post-training evaluation. We will now delve into detailed explanation of our scoring system.

The formulation of a scoring system for Vietnamese poetry engendered considerable complexity. Using the "luc bat" genre as an example, although it may be a familiar and easily comprehensible form for many Vietnamese, its governing rules are, in fact, notably intricate. The delineation of these rules is as follows:

$\bullet$ The word count of each line must alternate between 6 and 8, starting from 6.

$\bullet$ The 2\textsuperscript{nd}-4\textsuperscript{th}-6\textsuperscript{th} word of the 6-word line must be of even-uneven-even tone respectively (or vice versa).

$\bullet$ The 2\textsuperscript{nd}-4\textsuperscript{th}-6\textsuperscript{th}-8\textsuperscript{th} word of the 8-word line must be of even-uneven-even-even tone respectively (or vice versa, depending on the previous 6-word line). And in each 8-word line, the 6\textsuperscript{th} and 8\textsuperscript{th} word are of different accent.

$\bullet$ The 6\textsuperscript{th} word in a 6-word line must rhyme with the 6\textsuperscript{th} word in the ensuing 8-word line, as well as with the 8\textsuperscript{th} word in the previous 8-word line (if available).

Despite its complexity, with each genre having its own rules, we can distill the criteria for ascertaining the quality of a poem into three principals: length, rhyme and tone. And given a "luc bat" poem of n lines: 

\begin{equation}\text{poem} = l_0, l_1, ..., l_{n-1}\end{equation}

The scores can be calculated as:

\begin{equation}
\text{L} = \frac{1}{n}\sum_{i=0}^{\frac{n}{2}-1} \left(equal(\delta_{l_{2i}}, 6) + equal(\delta_{l_{2i+1}}, 8)\right)
\end{equation}

Where L is the length score of the poem's line-pairs and $\delta$ is the line's word count. For every line that is equal to the count, we add 1, and take the average after summation. As poems usually have even number of lines, if n is odd, we increase it by 1 as penalty. The tone score is defined as:

\begin{equation}
\text{T} = \frac{1}{n}\sum_{i=0}^{\frac{n}{2}-1} \left(\frac{match(l_{2i}, tone_6)}{3} + \frac{match(l_{2i+1}, tone_8)}{5}\right)
\end{equation}

Where 6-word lines must match the 3 tones, and for 8-word lines, 5 tones, defined in the "luc bat" rules above. Here, we add 1 for every tonal match then take the average. And the rhyme score can be calculated as:

\begin{equation}
\text{R} = \frac{2}{n}\sum_{i=0}^{\frac{n}{2}-1}\frac{rhyme(l_{2i-1}^8, l_{2i}^6, l_{2i+1}^6)}{t}
\end{equation}
\[
t = \begin{cases}
    2 & \text{if} \quad i = 0 \\
    3 & \text{otherwise}
\end{cases}
\]

Where the score of each pair (or triplet) is how many words rhyme according to the "luc bat" rules, divided by how many words are available. If no word rhymes then the score of that pair (or triplet) is $1/t$. Finally, we take the three scores and combine them according to the following formula:

\begin{equation}\text{score} = \frac{L}{10} + \frac{T \times 3}{10} + \frac{R \times 6}{10}\end{equation}

Here we give rhyme score higher weight than the rest so during the filtration step, rhymed poems are more likely to be selected. We then filter all poems in the dataset, only taking ones with the score of 0.9 or higher, which results in cutting the amount of samples by two third to over 50,000.

\section{Methodology}

In this section, we will discuss the specifics of our methodology. We will talk about our selection of models, their advantages and limitations. Followed by a detailed examination of the training process for poem generation.

\subsection{Architectures}
We leverages two prominent LLMs in its exploration of poem generation: GPT-3 and BLOOM. The selection of these models was guided by a careful consideration of their respective strengths and suitability for the research objectives.

\subsubsection{GPT-3} Developed by OpenAI, GPT-3 features a robust Transformer architecture and has undergone training on an extensive 45 Terabytes dataset encompassing both text and code. With various variants available in different sizes, the model showcases an impressive ability to comprehend and generate high-quality human-like language. For this particular experiment, we opted for the Davinci and Babbage variants. Davinci, the largest among the GPT-3 models with a parameter count of 175 billion, stands out as one of the most powerful and capable large language models (LLMs) currently available. However, its efficacy comes with a high finetuning cost of \$0.02 per 1,000 tokens (at the time of the assessment). In contrast, the Babbage variant, with 6 billion parameters, is significantly more cost-effective, priced at \$0.0004 per 1,000 tokens, making it an economical alternative, and the main testing variant of our choice.

\subsubsection{BLOOM} As an open-source offering from Hugging Face and BigScience, BLOOM is a collection of varying size models that has captured considerable attention within the LLM community. Similar to GPT-3, it employs a Transformer architecture and has been trained on a colossal dataset of 1.61 Terabytes of text, encompassing 46 languages and 13 programming languages. With 7.1 billion parameters, this BLOOM variant of choice is on par with Babbage, which minimizes the influence of size discrepancies on performance, allowing for a more nuanced investigation on poem generation.

However, the model's size still exceeded the capacity of our training environment. To address this computational bottleneck, we performed 8-bit quantization \cite{krishnamoorthi2018quantizing} to reduce the model's memory footprint by four, as well as LoRa adaptation \cite{hu2021lora} to partially freeze the model's parameters and only finetune the most significant ones. This optimization strategy enabled efficient training within our resource constraints while preserving model performance. Next, we will discuss the finetuning pipeline of our models.

\subsection{Poem Generation}
For the process of poem generation finetuning, we devise two downstream approaches: text-to-poem generation, and poem-to-poem generation.
\subsubsection{Text-to-poem generation}

\begin{figure}[htbp]
  \centering
  \includegraphics[width=0.4\textwidth]{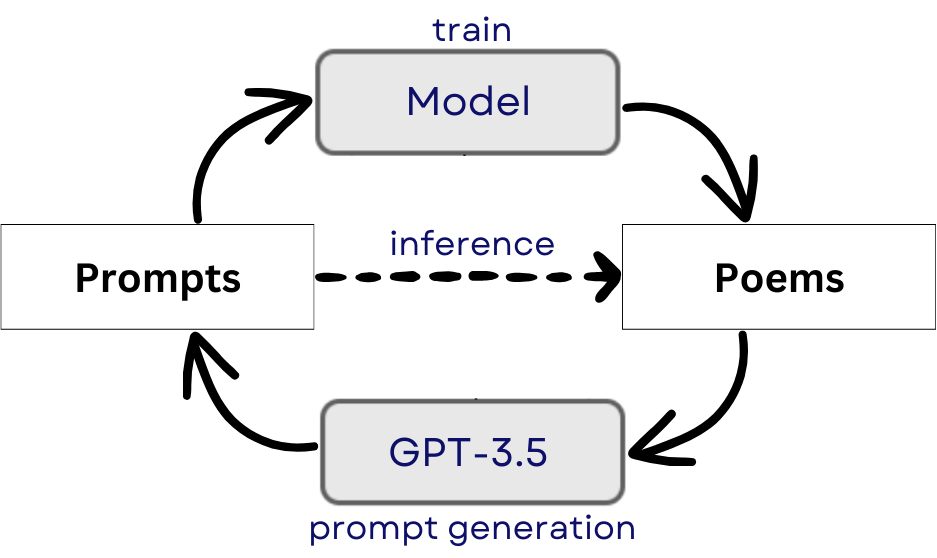} 
  \caption{The text-to-poem pipeline}
  \label{fig:t2p}
\end{figure}

Since the objective of this model is to let user input prompts of various length, context and requirements, we use the current GPT-3.5 to reverse generate prompts from poems, as seen in Figure \ref{fig:t2p}. Details of our prompt synthesization are as follows:

\begin{equation}
prompt = \mathcal{T}(X,Y,Z)
\end{equation}

In this context, $\mathcal{T}$ is the template used to guide the creation of a poem. The genre of the poem is represented by the variable $X$, while the topic is represented by $Y$. The sequence of keywords to be included in the poem is represented by $Z$. So the generated result could be like:

\[
\text{"Write a genre $X$ poem about $Y$, containing keywords $Z$"}
\]

Note that the actual generated prompts are in Vietnamese because we expect the model to be used in Vietnamese, for both input and output. After acquiring the prompt dataset, we finetuned GPT-3 (Davinci and Babbage) and BLOOM to generate poems from the prompts.

\subsubsection{Poem-to-poem generation}
\begin{figure}[htbp]
  \centering
  \includegraphics[width=0.4\textwidth]{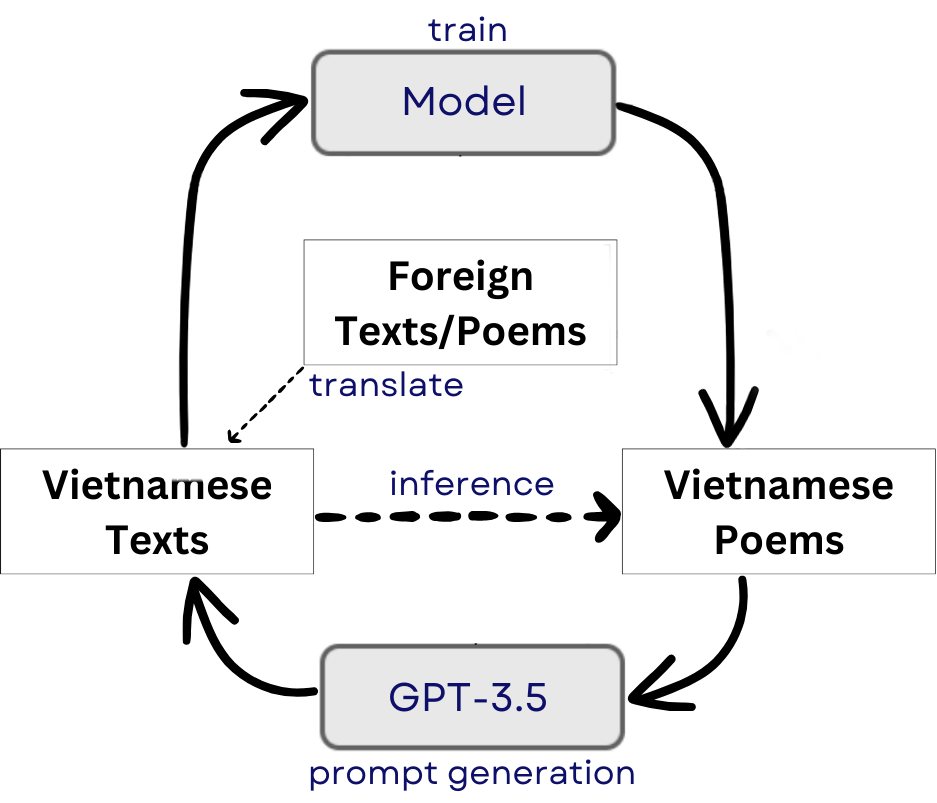} 
  \caption{The poem-to-poem pipeline}
  \label{fig:p2p}
\end{figure}

For this experimental approach, We created a new dataset, in which, we turned the poems into pure texts, paraphrased and used the resulted texts as input prompts, as illustrated in Figure \ref{fig:p2p}. This allows for the capturing of all context within the input to be use directly for generation, word by word. With this method, is possible to use any foreign piece of text, including foreign poems, preprocessed through a pipeline of translating to Vietnamese. For this method, we only train on the particular "luc bat" genre as it is the most popular genre of Vietnamese poems in our dataset.

\section{Performance Evaluation }
\subsection{Experiment setup}
Our GPT-3 models underwent finetuning via OpenAI's API, with a maximum allocation of \$18 (or one trial grant) assigned to each model. An additional \$18 was reserved for the prompt generation process using the ChatGPT API. This configuration facilitated the generation of a total of 49,958 samples for text-to-poem training, as well as 18,931 "luc bat" samples for poem-to-poem training, with 480 and 100 samples for testing respectively. It is noteworthy that not all training samples could be utilized in their entirety due to the monetary constraints of \$18 at the time. Consequently, the finetuning of the Davinci model for text-to-poem was carried out using only 500 samples of "luc bat" poems over 2 epochs (or 1,000 in total). In the case of the Babbage model, three approaches were implemented. For text-to-poem, the first model utilized all 49,958 samples of the dataset. The second model, on the other hand, only took 20,000 samples, serving as an investigation into the impact of sample size on performance, as well as to ensure fair comparison with BLOOM, the reason for which will be discussed later in this paragraph. Lastly, for poem-to-poem, the third model was finetuned with the full 18,931 samples. As for BLOOM, the finetuning would be performed on Google Colab, which, due to resource limitation, only allowed us to finetune one model with 20,000 samples.

To assess the efficacy of our models across various genres, we once again employed our established scoring system. We also compared our result against the zero-shot ChatGPT using the same testing data. Additionally, a distinct evaluation process, termed the "blind test," was conducted wherein genres were not disclosed. The creation of this test involved masking out the genre information in the prompts of the testing data. Subsequently, to identify the generated genre before scoring, we trained a genre classifier using BERT \cite{devlin2019bert}. This classifier processes data in the following formula: 
\begin{equation}
    genre = \text{Classifier}(\delta_{l_0}, \delta_{l_1}, ..., \delta_{l_{n-1}})
\end{equation}
    
Where the input is a string containing the word count $\delta$ of each line and demonstrated an accuracy of 99.7\%.

\subsection{Performance results}

\begin{table*}
  \centering
    \caption{Result comparison of models}
  \begin{tabular}{|c|c|c|c|c|c|c|}
    \hline
    \textbf{Models} & \textbf{Luc Bat}& \textbf{Blind}  & \textbf{7 Chu} & \textbf{8 Chu} & \textbf{5 Chu} & \textbf{4 Chu} \\
    \hline
    \multicolumn{7}{|c|}{text-to-poem} \\
    \hline
    ChatGPT (zero-shot)           & 0.440& 0.345&	0.292& 0.197&	0.284& 0.238 \\
    Davinci (1000 samples)    & 0.580&	-& -&	-&  -& -\\
    BLOOM (20k samples)              & 0.678& 0.596&	0.367& 0.279 &	0.480&  0.440\\
    Babbage (20k samples)              & 0.718&	-& -&	-&  -& -\\
    Babbage              & 0.805& 0.795&	0.661& 0.500&	0.382& 0.392\\
    \hline
    \multicolumn{7}{|c|}{poem-to-poem} \\
    \hline
    Babbage   & 0.781&	-& -&	-&  -& -\\
    \hline
  \end{tabular}

  \label{tab:result}
\end{table*}

\begin{figure*}[htbp]
  \centering
  \includegraphics[width=0.7\textwidth]{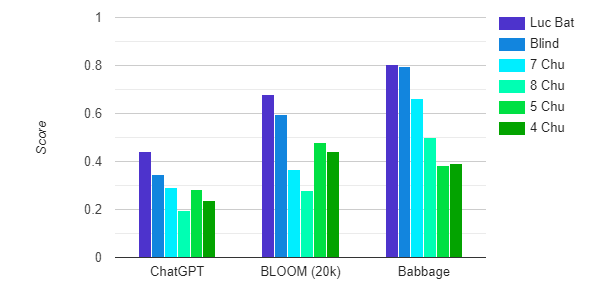} 
  \caption{Result comparison graph}
  \label{fig:result}
\end{figure*}

As depicted in Table \ref{tab:result} and Figure \ref{fig:result}, the optimal outcome is observed in the Babbage model finetuned on the entire dataset, yielding a score of 0.805 for "luc bat" and 0.795 for the blind test. This result aligns with expectations, considering the finetuning on the entire available dataset. Notably, given the larger sample count for the "luc bat" genre, the model exhibits a tendency to generate "luc bat" poems in cases where the genre is unspecified. Whereas for other genres, as their proportions in the dataset decrease, so do their scores.

For the BLOOM and Babbage models finetuned with 20,000 samples, sudden financial depletion prevented us from evaluating Babbage beyond "luc bat". A comparative analysis on this genre reveals that Babbage outperforms BLOOM slightly in the "luc bat" genre, with the score of 0.718 and 0.678 respectively, while having lower parameter count. With GPT-3 technology being proprietary, we can not draw any insight from the architectural differences yet, but it is possible that such architecture, combining with more extensive training from the OpenAI team, possesses more comprehensive understanding for downstream tasks. Furthermore, the significant reduction in sample size from 50,000 to 20,000 has negatively impacted Babbage's scores, which, interestingly, in the case of the latter Babbage model, the scores for the "4 chu" and "5 chu" genres appear unusually high despite fewer samples. This anomaly is attributed to underfitting, resulting in the model repeating and duplicating lines, thereby elevating the tone and rhyme score. For the Davinci model, constrained by a limited sample size of 500 over two epochs, it achieves a comparatively low score of 0.58 while costing the same amount of money for finetuning as Babbage. But despite the aforementioned variations, all their results surpass those obtained using the zero-shot ChatGPT, which lacks an understanding of the specified genre and consistently generates poems of inconsistent word counts with no regard for tonal or rhyme rules.

As for the poem-to-poem generation, we obtain a score of 0.781. This figure is slightly below the outcome of the preceding Babbage experiment, despite maintaining an similar number of "luc bat" samples. The discrepancy arises due to the heightened complexity of the input prompts, now necessitating the model not only to integrate them into the generation but also to paraphrase them into synonyms that align with the prescribed tone and rhyme rules. However, the diminished score does not inherently denote an inferior approach. On the contrary, it confers complete control over the generated content, a capability that the previous technique struggles slightly. Furthermore, this methodology introduces a novel dimension in the form of cross-language poem-to-poem translation. With an input text or poem from a foreign language, it becomes conceivable to generate a corresponding and correctly formulated Vietnamese poem.

\section{Conclusion}

In summary, our project has achieved a commendable level of success. Our initial objective was to construct a model capable of creatively and accurately generating Vietnamese poems, drawing inspiration from the recent accomplishments in language models. The outcome surpasses the mere creation of a Vietnamese poem generator; we introduced the novel pipeline for poem-to-poem generation, unbound by language barriers. We have also introduced a comprehensive poem scoring method, customizable to individual user requirements. While we aspired to compare our models with Tuan Nguyen et al.'s SP-GPT2, the unavailability of functional code from their repository prevented a direct comparison. As we contemplate further enhancements to this project, one viable avenue involves augmenting the dataset size by artificially generating new data using the model and subsequently filtering for high-quality samples. However, the most apparent solution lies in training the entire dataset on the more potent Davinci or even better future iterations. While this proposition may pose a costly challenge for a smaller team, it remains well within the realm of feasibility for larger entities such as corporations and research institutions.

\balance
\bibliographystyle{IEEEtran}


\end{document}